\documentclass[11pt]{article}

\usepackage[margin=1in]{geometry}
\usepackage{graphicx}
\usepackage{amsmath,amssymb}
\usepackage{booktabs}
\usepackage{microtype}
\usepackage{hyperref}
\usepackage{xcolor}
\usepackage{caption}
\usepackage{subcaption}
\usepackage{authblk}

\hypersetup{colorlinks=true,linkcolor=blue!60!black,citecolor=blue!60!black,urlcolor=blue!60!black}

\title{SciDraw-6K: A Multilingual Scientific Illustration Dataset Generated by Google Gemini}
\author[1]{Davie Chen\thanks{\texttt{daviechen@bsu.edu.pl} \quad ORCID: \href{https://orcid.org/0009-0001-4819-2828}{0009-0001-4819-2828}}}
\affil[1]{University of Arts in Pozna\'n}
\date{\today}

\begin{document}
\maketitle

\begin{abstract}
We present SciDraw-6K, a curated dataset of 6{,}291 scientific illustrations
synthesized by Google Gemini image-generation models, each paired with prompts
in eleven languages (English, Chinese (Simplified and Traditional), Japanese,
Korean, German, French, Spanish, Brazilian Portuguese, Italian, and Russian).
Images span eight broad scientific categories---biomedical, chemistry,
materials, electronics, environment, AI systems, physics, and a long
``other''~tail---and are produced primarily by the
\texttt{gemini-2.5-flash-image} and \texttt{gemini-3-pro-image-preview} model
families. In contrast to general-purpose text-to-image (T2I) corpora that
dominate the literature, SciDraw-6K is purpose-built for the scientific
illustration genre: schematic diagrams, mechanism figures, table-of-contents
graphics, and conceptual posters. We describe the construction pipeline,
report dataset statistics, and document its use as the substrate of
\href{https://sci-draw.com}{sci-draw.com}, a public scientific drawing
service. The dataset is released to support multilingual T2I research,
domain-adapted diffusion fine-tuning, and prompt-engineering studies for
scientific visualization. The dataset is available at
\url{https://huggingface.co/datasets/SciDrawAI/SciDraw-6K} and archived at
\url{https://doi.org/10.5281/zenodo.19642870}.

\end{abstract}

\section{Introduction}

Modern text-to-image (T2I) systems have rapidly improved on photorealistic
and artistic generation, but \emph{scientific illustration}---the schematic,
diagrammatic, and conceptual imagery used in research papers, talks, and
textbooks---remains a comparatively underserved regime. Scientific figures
demand a distinctive visual grammar (clean labels, abstracted geometry,
minimal photorealism, well-balanced layouts) and a domain-aware semantics
(correct molecular topology, realistic device cross-sections, biologically
plausible cell anatomies, etc.). Existing T2I corpora such as
LAION-5B~\cite{schuhmann2022laion5b}, JourneyDB~\cite{pan2023journeydb} and
DiffusionDB~\cite{wang2023diffusiondb} are dominated by stylistic, aesthetic
or photorealistic content scraped from the open web; explicitly scientific
subsets are small or implicit.

In parallel, frontier proprietary models---in particular Google's
Gemini~\cite{gemini2024report} image-generation family---have demonstrated
strong zero-shot capabilities on scientific schematics, suggesting an
opportunity to bootstrap a public dataset by treating a strong closed model
as a synthetic data engine, an approach popularized by
self-instruct-style pipelines~\cite{wang2023selfinstruct}.

We contribute \textbf{SciDraw-6K}, a dataset of $6{,}291$ Gemini-generated
scientific illustrations released together with this report. Each image
ships with eleven aligned multilingual prompts and a coarse subject-matter
category. Concretely, our contributions are:
\begin{enumerate}
  \item A reproducible construction pipeline that combines a
        domain-specific prompt taxonomy, Gemini image generation, automated
        translation into ten target languages, and lightweight quality
        control.
  \item Per-image metadata including raw and release categories, model
        identifier when recoverable from generation logs, generation type,
        SHA-256 hash, and 11-language prompts, enabling multilingual T2I
        research while preserving the distinction between the original
        long-tail taxonomy and the eight-class release taxonomy.
  \item A descriptive statistical analysis of the dataset (Section~4) and a
        case study of how SciDraw-6K powers
        \href{https://sci-draw.com}{sci-draw.com}, a public scientific
        drawing service (Section~5).
\end{enumerate}

We position SciDraw-6K as a complement to general T2I datasets: it is small
in scale but high in domain density and language coverage, and is intended
to support fine-tuning, retrieval-augmented generation, multilingual prompt
engineering, and benchmarking for scientific visualization.

\paragraph{Availability.}
SciDraw-6K~\cite{chen2026scidraw6k} is hosted on the Hugging Face Hub at
\url{https://huggingface.co/datasets/SciDrawAI/SciDraw-6K} and archived
with a permanent DOI at
\url{https://doi.org/10.5281/zenodo.19642870}. The dataset is released
under CC~BY~4.0.

\section{Related Work}

\paragraph{Large-scale T2I corpora.}
LAION-5B~\cite{schuhmann2022laion5b} pioneered web-scale image--text
datasets and underpinned many open diffusion
models~\cite{rombach2022stable}. JourneyDB~\cite{pan2023journeydb} curated
millions of Midjourney generations with human-style prompts;
DiffusionDB~\cite{wang2023diffusiondb} similarly collected
Stable~Diffusion outputs with their textual prompts. These corpora are
broad and stylistically diverse but contain very little explicitly
scientific content; their captions are typically aesthetic,
single-language (English-dominant), and not aligned across languages.

\paragraph{Scientific visualization datasets.}
Datasets such as SciCap~\cite{hsu2021scicap} and
FigureQA~\cite{kahou2017figureqa} target chart-style figures with
machine-readable structure rather than free-form schematic illustrations.
Datasets of biomedical figures (e.g.\ ROCO~\cite{pelka2018roco}) focus on
radiology rather than synthetic schematics. To our knowledge there is no
public dataset that combines (i)~scientific-illustration style,
(ii)~broad multi-disciplinary coverage and (iii)~aligned multilingual
captions.

\paragraph{Self-instruct and synthetic data.}
Treating a strong closed model as a teacher to generate training data is
now standard practice in instruction tuning~\cite{wang2023selfinstruct}
and increasingly in vision~\cite{bai2024coyo}. SciDraw-6K applies the
same recipe to scientific illustration: a single high-quality T2I
model (Gemini) is used both to author the images and, in concert with
LLM-driven translation, to populate the multilingual side of each
example.

\paragraph{Multilingual T2I.}
Most public T2I checkpoints are trained on English-heavy data and degrade
on non-English prompts. Recent efforts such as
AltDiffusion~\cite{ye2023altdiffusion} and PaLI~\cite{chen2023pali}
emphasize the value of aligned multilingual supervision. SciDraw-6K
provides eleven aligned languages per image, with full coverage in each
column, and is therefore directly applicable to multilingual prompt
robustness studies.

\begin{table}[t]
\centering
\small
\begin{tabular}{lccccc}
\toprule
Dataset & \#images & sci.\ focus & multilingual & source model \\
\midrule
LAION-5B           & $5\cdot10^9$ & no  & partial & web                \\
JourneyDB          & $4\cdot10^6$ & no  & no      & Midjourney         \\
DiffusionDB        & $1.4\cdot10^7$ & no & no    & Stable Diffusion   \\
SciCap             & $4\cdot10^5$ & yes (charts) & no & paper figures \\
\textbf{SciDraw-6K (ours)} & $6.3\cdot10^3$ & \textbf{yes} & \textbf{11 langs} & Gemini \\
\bottomrule
\end{tabular}
\caption{Comparison with representative public T2I datasets. SciDraw-6K
trades scale for domain density and language coverage.}
\end{table}

\section{Dataset Construction}

\subsection{Source model}
All images in SciDraw-6K were synthesized by Google Gemini image-generation
endpoints. The dominant model variants are
\texttt{gemini-2.5-flash-image} and \texttt{gemini-3-pro-image-preview},
with a small tail produced by
\texttt{gemini-3.1-flash-image-preview}. Generation was performed via the
official Gemini API at default safety and resolution settings; per-image
metadata records the exact model identifier, allowing downstream
stratification by model generation.

\subsection{Prompt taxonomy}
We developed a coarse eight-category taxonomy specifically targeted at
scientific illustration:
\textbf{biomedical}, \textbf{chemistry}, \textbf{materials},
\textbf{electronics}, \textbf{environment}, \textbf{ai\_system},
\textbf{physics}, and a residual \textbf{other} bucket that absorbs
long-tail areas (robotics, economics, mathematics, civil engineering,
geoscience, \dots). Internally, the authoring workflow uses a slightly
finer raw label space; the public release therefore exposes both
\texttt{raw\_category} and the normalized \texttt{release\_category}
used in this paper. For each category we curated a small library of prompt
templates targeting the most common scientific-figure modalities:
\emph{table-of-contents (TOC) graphical abstracts}, \emph{mechanism
diagrams}, \emph{device cross-sections}, \emph{conceptual schematics}, and
\emph{summary posters}. Templates leave free slots for the user-supplied
research topic, which is composed into a concrete prompt by the
authoring application before being sent to Gemini.

\subsection{Multilingual translation}
Each image is paired with prompts in eleven languages: English (en),
Simplified Chinese (zh), Japanese (ja), Korean (ko), German (de), French
(fr), Spanish (es), Brazilian Portuguese (pt\_br), Traditional Chinese
(zh\_tw), Italian (it), and Russian (ru). Translations are generated by an
LLM-based pipeline conditioned on the original prompt. After translation,
all 11 columns are populated for 100\% of approved rows
(see Section~4), giving a fully aligned parallel text resource in the
sense of field coverage. We stress, however, that full coverage should not
be read as a guarantee of native-speaker fluency or error-free encoding;
the release pipeline includes automated checks for suspicious mojibake
patterns, and multilingual quality still warrants manual auditing.

\subsection{Quality control and anonymization}
A lightweight curation pass marks each image as approved or rejected based
on (i)~obvious generation failure (e.g.\ blank or corrupted output),
(ii)~off-topic content, and (iii)~policy violations. Only approved rows
are included in the released dataset. To eliminate any link to the
authoring application's user accounts, we strip user identifiers,
conversation identifiers, and message identifiers prior to export. The
released metadata contains only: a stable image~ID, the public image~URL,
the file extension, raw and release category labels, the eleven
multilingual prompts, the Gemini model identifier when available from the
generation table, the generation type, the creation timestamp, and the
SHA-256 hash of the downloaded image bytes. We additionally emit a machine
readable validation report that records missing provenance fields and
suspicious multilingual strings detected during export.

\subsection{Release format}
The dataset is distributed as (i)~a JSON-Lines file
(\texttt{metadata.with\_hash.jsonl}) and an equivalent Apache~Parquet file
(\texttt{metadata.parquet}); (ii)~the images themselves, organized as
\texttt{images/<release\_category>/<id>.<ext>}; (iii)~a validation summary
(\texttt{metadata.validation.json}); and (iv)~a prompt-grouped
train/validation/test split file (\texttt{splits.json}) intended to reduce
near-duplicate leakage between evaluation partitions. A reproducible
download script and the construction pipeline are released alongside the
data. The dataset is hosted on Hugging Face
(\texttt{SciDrawAI/SciDraw-6K}) with a permanent Zenodo DOI
(\texttt{10.5281/zenodo.19642870}) for citation~\cite{chen2026scidraw6k}.

\section{Dataset Statistics}

SciDraw-6K contains $6{,}291$ approved images spanning the period
2026-01 through 2026-04. We summarize the most informative dimensions
below; all figures are produced from \texttt{metadata.with\_hash.jsonl}
by \texttt{scripts/05\_compute\_stats.py}.

\begin{figure}[t]
\centering
\includegraphics[width=0.85\linewidth]{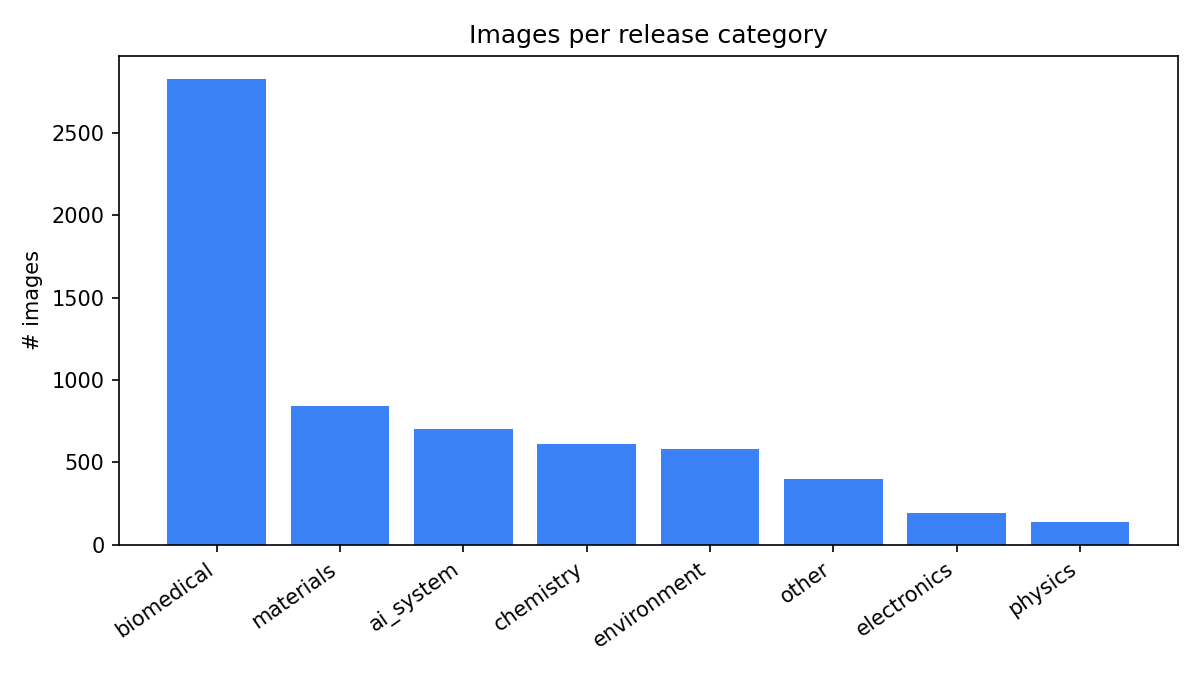}
\caption{Number of images per category. The biomedical category
dominates, with a long thin tail of less common disciplines absorbed
into ``other''.}
\label{fig:cat}
\end{figure}

\paragraph{Category distribution.}
Figure~\ref{fig:cat} shows the per-category counts. Biomedical
illustrations dominate with roughly 45\% of all images ($2{,}827/6{,}291$),
followed by materials science ($841$), AI~systems ($705$), chemistry
($609$), and environmental science ($581$). Electronics ($190$) and
physics ($139$) are comparatively small, while the residual ``other''
bucket contains a very sparse long tail of raw labels including robotics,
mathematics, economics, civil engineering, and geosciences. The skew
reflects real-world demand on the authoring service rather than an explicit
sampling target, and we recommend re-weighting or stratified sampling for
downstream training if balanced category coverage is desired.

\begin{figure}[t]
\centering
\includegraphics[width=0.85\linewidth]{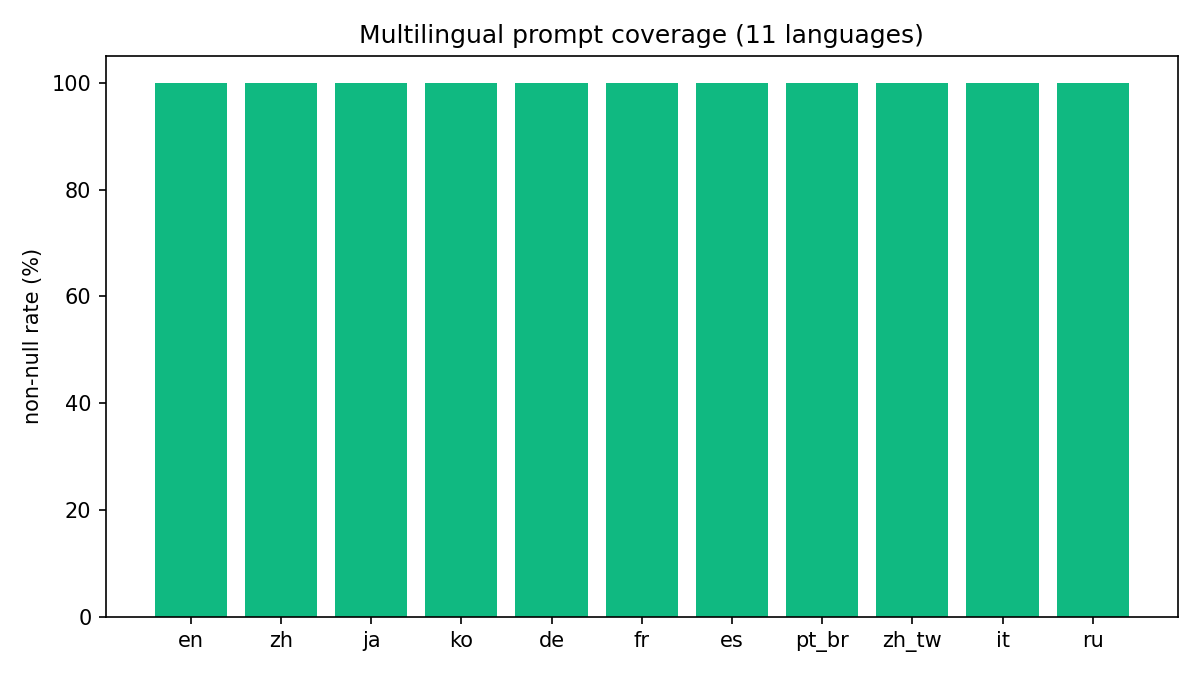}
\caption{Per-language non-null rate of prompt fields. All eleven
languages are populated for 100\% of released images.}
\label{fig:lang}
\end{figure}

\paragraph{Multilingual coverage.}
Figure~\ref{fig:lang} confirms that all eleven prompt languages are
present for every approved row: each language column has 100\%
non-null coverage, yielding a fully aligned $6{,}291 \times 11$ parallel
text matrix. This is a coverage claim only: it does not by itself
establish translation adequacy, native-speaker naturalness, or encoding
cleanliness. In the released pipeline we therefore pair coverage reporting
with an automated validation report that flags suspicious multilingual
strings for follow-up inspection.

\begin{figure}[t]
\centering
\includegraphics[width=0.85\linewidth]{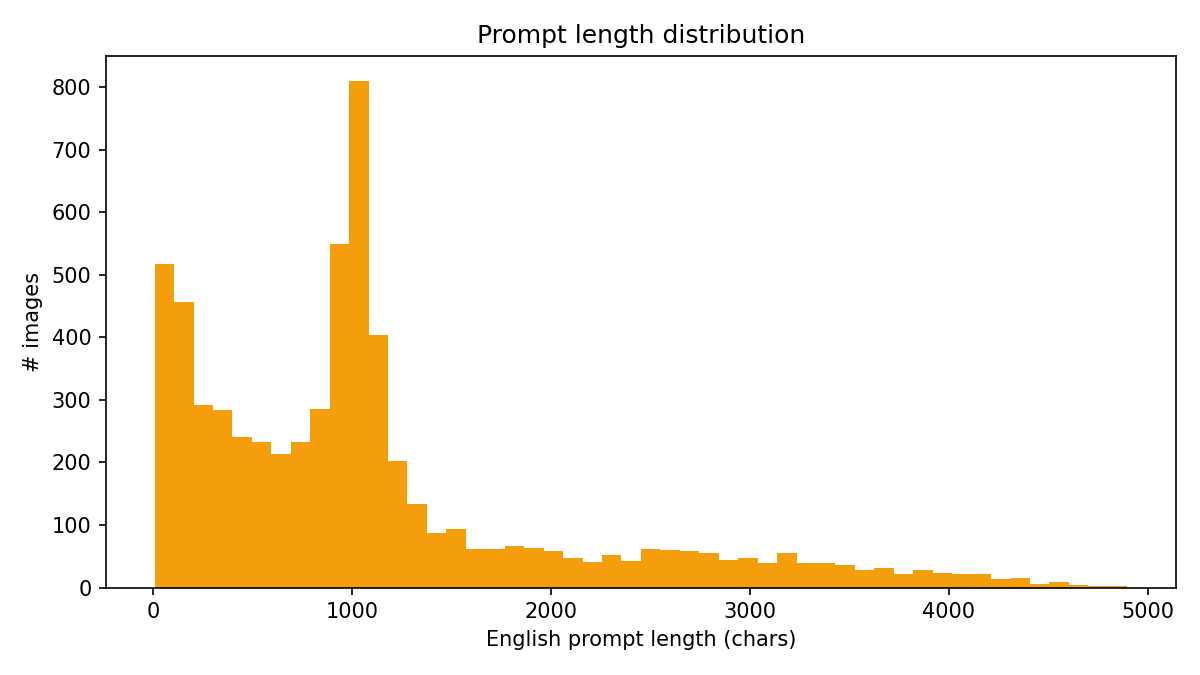}
\caption{English prompt length distribution (characters).}
\label{fig:plen}
\end{figure}

\paragraph{Prompt length.}
Figure~\ref{fig:plen} reports the distribution of English prompt length
in characters. Prompts are typically several hundred characters long,
reflecting the structured, template-driven authoring process and the
relatively dense semantics required for scientific schematics; the
distribution exhibits a long right tail of highly detailed prompts.

\begin{figure}[t]
\centering
\includegraphics[width=0.85\linewidth]{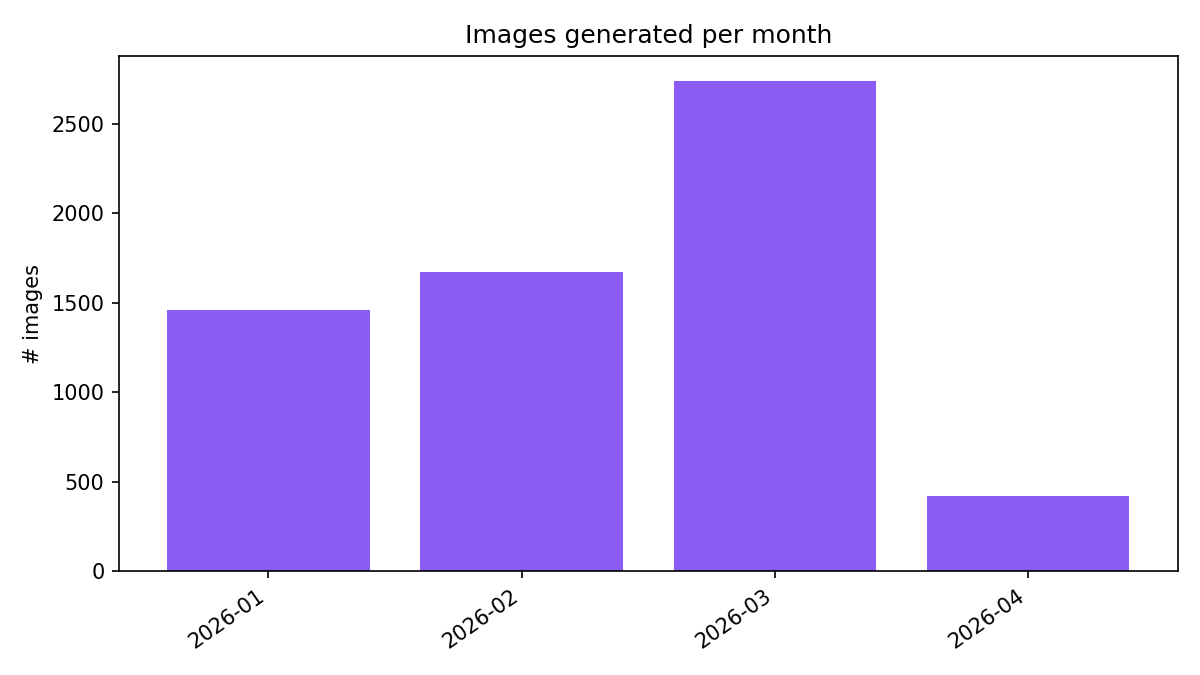}
\caption{Number of images generated per month.}
\label{fig:time}
\end{figure}

\paragraph{Temporal distribution.}
Figure~\ref{fig:time} shows monthly generation counts. Volume grows
over time as both the authoring service and Gemini's image-generation
capabilities matured.

\begin{figure}[t]
\centering
\includegraphics[width=0.85\linewidth]{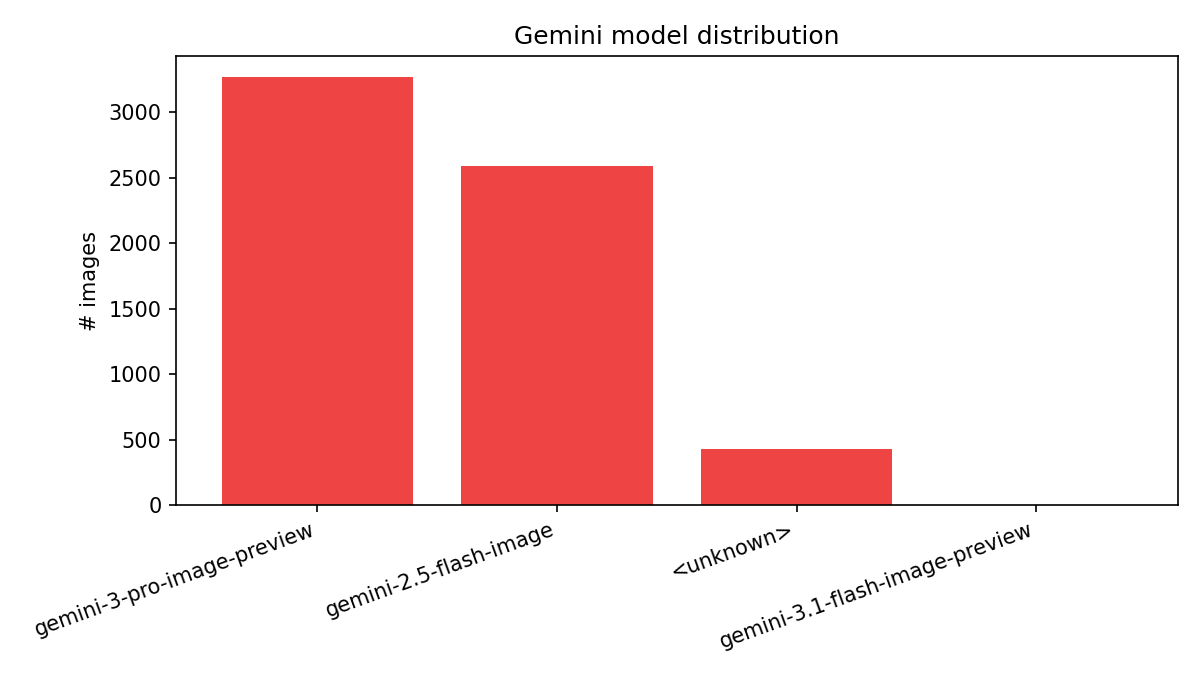}
\caption{Gemini source-model distribution across approved images.}
\label{fig:model}
\end{figure}

\paragraph{Source-model distribution.}
Figure~\ref{fig:model} breaks down approved images by Gemini model
identifier. The bulk of the dataset is split between
\texttt{gemini-3-pro-image-preview} and \texttt{gemini-2.5-flash-image},
with a very small \texttt{gemini-3.1-flash-image-preview} contribution.
Importantly, not all approved rows can be linked back to a successful
Gemini generation log entry: $430$ rows (about 6.8\%) have unknown model
and generation-type provenance in the public export. We retain these rows
because the image artifacts and prompts are valid, but analyses stratified
by source model should either exclude or separately report this unknown
slice. This split lets researchers study how source-model identity affects
downstream behaviour, e.g.\ when fine-tuning open diffusion models on
subsets stratified by teacher.

\section{Application: \texttt{sci-draw.com}}

To demonstrate the practical value of SciDraw-6K, we describe its use as
the substrate of \href{https://sci-draw.com}{sci-draw.com}, a public
scientific drawing service. The site exposes Gemini-style image
generation through a domain-specialized interface aimed at researchers,
educators, and science communicators.

\paragraph{System overview.}
sci-draw.com is a Next.js web application with a PostgreSQL backend.
It provides four generation modes---\emph{prompt-to-image},
\emph{sketch-to-image}, \emph{image-to-image}, and a \emph{custom}
mode---wrapped in a credit-based usage system. Each session is logged
to a structured generation history, which we periodically curate into
the gallery from which SciDraw-6K is constructed.

\paragraph{How SciDraw-6K is used.}
The dataset plays three concrete roles in the live service:
\begin{itemize}
  \item \textbf{Template seeding.} Editors browse SciDraw-6K to
        identify high-quality prompts and reuse them as one-click
        templates organized by category. The aligned 11-language
        prompts let new users start in their preferred language.
  \item \textbf{Few-shot prompt rewriting.} When a user submits a
        terse natural-language request, the application retrieves
        the nearest neighbours from SciDraw-6K (by category and
        embedding) and uses their (longer, well-structured) prompts
        as in-context exemplars to rewrite the user's request before
        sending it to Gemini.
  \item \textbf{Internal evaluation.} A held-out slice of the dataset
        is used as a regression suite when we upgrade the underlying
        Gemini model: we re-render its prompts on the new model and
        spot-check that scientific quality (label legibility,
        topology correctness, layout coherence) does not regress.
\end{itemize}

\paragraph{Discussion.}
The key insight is that a small but high-quality, multi-disciplinary,
multilingual dataset is sufficient to bootstrap a useful production
service in the scientific-illustration genre. We do not train a
diffusion model from scratch on SciDraw-6K; instead we use the dataset
as templates, retrieval pool, and regression suite around a strong
proprietary base model. This is a deliberately lightweight pattern that
we believe is easy for other research groups to replicate in adjacent
domains.

\section{Limitations and Ethical Considerations}

\paragraph{Single-source bias.}
All images are generated by a single proprietary model family
(Google Gemini). Stylistic, compositional, and semantic biases of that
model are therefore baked into SciDraw-6K. Models trained or evaluated
exclusively on this dataset risk overfitting to Gemini-specific
artefacts. We recommend treating the dataset as a complement to, not a
replacement for, broader T2I corpora.

\paragraph{Category imbalance.}
The biomedical category dominates the dataset, while several scientific
disciplines (mathematics, civil engineering, geosciences) are
represented by fewer than ten images each. Downstream training jobs
should oversample minority classes or treat the long tail with care. More
generally, SciDraw-6K should be interpreted as a demand-shaped production
corpus rather than a balanced survey of scientific illustration.

\paragraph{English-anchored multilinguality.}
Although prompts are released in eleven languages, the original
authoring interface and the LLM-driven translation pipeline are both
English-anchored. The non-English columns are translation-quality
proxies of the English original, not independent native-language
captions. They are useful for multilingual prompt-robustness studies
but should not be interpreted as a balanced multilingual user corpus.
Automated export checks can catch some encoding anomalies, but they do not
replace human review for fluency or domain terminology.

\paragraph{Incomplete provenance fields.}
Not every approved gallery row can be cleanly rejoined to a successful
Gemini generation log entry, leaving a non-trivial unknown slice in the
released \texttt{gemini\_model} and \texttt{generation\_type} columns.
This does not invalidate the image--prompt pairs, but it does limit any
analysis that attempts to compare source-model families at high precision.

\paragraph{Template-heavy prompts.}
The dataset is optimized for practical scientific-figure generation rather
than caption diversity. Many prompts follow reusable authoring templates,
and some near-duplicates differ only in inserted topics or minor phrasing.
We mitigate evaluation leakage by releasing prompt-grouped splits, but
users should still avoid treating the corpus as a naturally occurring
caption distribution.

\paragraph{Model and policy compliance.}
All images were generated under Google's Gemini terms of service, and
the release is intended to remain consistent with those terms at the time
of publication. Because platform policies can evolve, downstream users
should independently verify current licensing and redistribution
constraints before republishing mirrors or derived artifacts. We have not
redistributed any end-user account information; all per-user identifiers
are removed prior to export.

\paragraph{Potential harms.}
Synthetic scientific imagery can in principle be misused to fabricate
plausible-looking but incorrect figures. We discourage use of
SciDraw-6K imagery as ground-truth scientific evidence; the dataset is
intended for visualization, education, and ML research purposes.

\section{Conclusion}

We have introduced SciDraw-6K, a small but high-density dataset of
$6{,}291$ Gemini-generated scientific illustrations with eleven aligned
multilingual prompts and an eight-category subject taxonomy, and
described its use as the substrate of the
\href{https://sci-draw.com}{sci-draw.com} service. We hope SciDraw-6K
proves useful for multilingual T2I research, domain-adapted fine-tuning
of diffusion models, retrieval-augmented prompt engineering, and the
broader study of how frontier image-generation models behave on the
specialized visual grammar of science.

\bibliographystyle{plain}
\bibliography{references}

@inproceedings{schuhmann2022laion5b,
  title={{LAION-5B}: An Open Large-Scale Dataset for Training Next Generation Image-Text Models},
  author={Schuhmann, Christoph and Beaumont, Romain and Vencu, Richard and Gordon, Cade and Wightman, Ross and Cherti, Mehdi and Coombes, Theo and Katta, Aarush and Mullis, Clayton and Wortsman, Mitchell and others},
  booktitle={NeurIPS Datasets and Benchmarks},
  year={2022}
}

@inproceedings{pan2023journeydb,
  title={{JourneyDB}: A Benchmark for Generative Image Understanding},
  author={Pan, Junting and Sun, Keqiang and Ge, Yuying and Li, Hao and Duan, Haodong and Wu, Xiaoshi and Zhang, Renrui and Zhou, Aojun and Qin, Zipeng and Wang, Yi and others},
  booktitle={NeurIPS},
  year={2023}
}

@article{wang2023diffusiondb,
  title={{DiffusionDB}: A Large-Scale Prompt Gallery Dataset for Text-to-Image Generative Models},
  author={Wang, Zijie J and Montoya, Evan and Munechika, David and Yang, Haoyang and Hoover, Benjamin and Chau, Duen Horng},
  journal={ACL},
  year={2023}
}

@article{wang2023selfinstruct,
  title={Self-Instruct: Aligning Language Model with Self Generated Instructions},
  author={Wang, Yizhong and Kordi, Yeganeh and Mishra, Swaroop and Liu, Alisa and Smith, Noah A and Khashabi, Daniel and Hajishirzi, Hannaneh},
  journal={ACL},
  year={2023}
}

@techreport{gemini2024report,
  title={Gemini: A Family of Highly Capable Multimodal Models},
  author={{Google DeepMind}},
  institution={Google},
  year={2024}
}

@inproceedings{rombach2022stable,
  title={High-Resolution Image Synthesis with Latent Diffusion Models},
  author={Rombach, Robin and Blattmann, Andreas and Lorenz, Dominik and Esser, Patrick and Ommer, Bj{\"o}rn},
  booktitle={CVPR},
  year={2022}
}

@inproceedings{hsu2021scicap,
  title={{SciCap}: Generating Captions for Scientific Figures},
  author={Hsu, Ting-Yao and Giles, C. Lee and Huang, Ting-Hao K.},
  booktitle={Findings of EMNLP},
  year={2021}
}

@article{kahou2017figureqa,
  title={{FigureQA}: An Annotated Figure Dataset for Visual Reasoning},
  author={Kahou, Samira Ebrahimi and Michalski, Vincent and Atkinson, Adam and K{\'a}d{\'a}r, {\'A}kos and Trischler, Adam and Bengio, Yoshua},
  journal={ICLR Workshop},
  year={2018}
}

@article{pelka2018roco,
  title={Radiology Objects in Context ({ROCO}): A Multimodal Image-Dataset},
  author={Pelka, Obioma and Koitka, Sven and R{\"u}ckert, Johannes and Nensa, Felix and Friedrich, Christoph M.},
  journal={MICCAI Workshop},
  year={2018}
}

@article{ye2023altdiffusion,
  title={{AltDiffusion}: A Multilingual Text-to-Image Diffusion Model},
  author={Ye, Fulong and Liu, Guang and Wu, Xinya and Wu, Lei},
  journal={arXiv preprint arXiv:2308.09991},
  year={2023}
}

@article{chen2023pali,
  title={{PaLI}: A Jointly-Scaled Multilingual Language-Image Model},
  author={Chen, Xi and Wang, Xiao and Changpinyo, Soravit and Piergiovanni, AJ and Padlewski, Piotr and Salz, Daniel and Goodman, Sebastian and Grycner, Adam and Mustafa, Basil and Beyer, Lucas and others},
  journal={ICLR},
  year={2023}
}

@misc{bai2024coyo,
  title={Synthetic Vision Datasets from Frontier Generative Models},
  author={Bai, Yuntao and others},
  year={2024},
  note={Survey reference}
}

@misc{chen2026scidraw6k,
  author       = {Chen, Davie},
  title        = {{SciDraw-6K}: A Multilingual Scientific Illustration Dataset Generated by {Google Gemini}},
  year         = {2026},
  publisher    = {Zenodo},
  doi          = {10.5281/zenodo.19642870},
  howpublished = {Zenodo},
  note         = {DOI: \href{https://doi.org/10.5281/zenodo.19642870}{10.5281/zenodo.19642870}},
  url          = {https://doi.org/10.5281/zenodo.19642870}
}

\end{document}